\documentclass[conference]{IEEEtran}

\usepackage{float} 
\usepackage{booktabs} 
\usepackage[utf8]{inputenc} 
\usepackage{amsfonts}
\usepackage{algorithm}
\usepackage{algpseudocode}
\usepackage{enumitem}

\usepackage{cite}
\usepackage{amsmath,amssymb,amsfonts}
\usepackage{textcomp}
\usepackage{xcolor}
\def\BibTeX{{\rm B\kern-.05em{\sc i\kern-.025em b}\kern-.08em
    T\kern-.1667em\lower.7ex\hbox{E}\kern-.125emX}}
\begin{document}

\title{Online Parallel Portfolio Selection with Heterogeneous Island Model}

\author{\IEEEauthorblockN{Štěpán Balcar}
\IEEEauthorblockA{\textit{Charles University} \\
\textit{Faculty of Mathematics and Physics}\\
Prague, Czech Republic\\
Stepan.Balcar@mff.cuni.cz}
\and
\IEEEauthorblockN{Martin Pilát}
\IEEEauthorblockA{\textit{Charles University} \\
\textit{Faculty of Mathematics and Physics}\\
Prague, Czech Republic\\
Martin.Pilat@mff.cuni.cz}
}

\maketitle

\begin{abstract}
We present an online parallel portfolio selection algorithm based on the island model commonly used for parallelization of evolutionary algorithms. In our case each of the islands runs a different optimization algorithm. The distributed computation is managed by a central planner which periodically changes the running methods during the execution of the algorithm -- less successful methods are removed while new instances of more successful methods are added.

We compare different types of planners in the heterogeneous island model among themselves and also to the traditional homogeneous model on a wide set of problems. The tests include experiments with different representations of the individuals and different duration of fitness function evaluations. The results show that heterogeneous models are a more general and universal computational tool compared to homogeneous models.
\end{abstract}


\section{Introduction}
In the past decades many different optimization methods have been proposed to solve various types of problems, including hill-climbing \cite{colbourn1985algorithms}, random search \cite{Rastrigin63},
simulated annealing \cite{Kirkpatrick83}, tabu search \cite{Glover86}, evolutionary algorithms \cite{holland75}, and differential evolution \cite{Price:2005:DEP:1121631}. The methods are based on various paradigms and some of them are better for one problem, while others are better for another. In fact, the most suitable method can even be different in different phases of the optimization. In the beginning of the optimization run, very often methods that are based on the exploration are preferred, while in the later phases, exploitation of already found areas of the search space may be more beneficial.  

The algorithm selection problem has been defined by Rice several decades ago as the problem of selecting the best algorithm for each problem instance from a set of instances such that the overall optimization cost is minimized~\cite{rice1976algorithm}. Similar types of problems are also solved in the area of machine learning, with the goal to select the best machine-learning method for a given dataset. In this case, the problem is called meta-learning~\cite{Brazdil:2008:MAD:1507541}.

In the recent years, computers with multiple CPU cores have become more common, and that brings an alternative option, how to deal with the problem of algorithm selection. In such a case the so called portfolio algorithms~\cite{GOMES200143} can be used. The portfolio algorithms run multiple (more or less) independent instances of different algorithms in parallel (or sequentially on a single CPU) with the goal to find better solution. Sometimes the instances differ only in random initialization of the algorithm parameters. 
Quite recently, Lindenauer \emph{et al.}~\cite{lindauer2015sequential} studied the problem of portfolio selection as an extension of the problem of algorithm selection.

In this paper, we combine some of the ideas mentioned above and provide in essence an online algorithm for parallel portfolio selection. The algorithm is based on a heterogeneous island model, which is derived from the homogeneous island model commonly used for the parallelization of evolutionary algorithms. The island model~\cite{kacprzyk2015springer} is based on the idea of the life of several isolated populations on different islands evolving in parallel. The islands cooperate with each other in the computation by exchanging the individuals from local populations, thus accelerating the convergence.

In homogeneous island models, each of the islands runs the same algorithm with the same settings, while in the heterogeneous model, the methods used on each island are somehow different. For example, Gong and Fukunaga \cite{Gong11} proposed a heterogeneous model where each island runs an evolutionary algorithm with different settings as an alternative adaptive approach to the manual parameter setting. Pilát and Neruda~\cite{DBLP:conf/cec/2010} introduced a heterogeneous models in the context of multi-objective optimization. Some of the islands run multi-objective algorithm, while the others run only a single-objective one. The single-objective islands help to improve the solutions of the multi-objective problem. The heterogeneity in this case actually serves as a way of creating a new hybrid evolutionary algorithm. 

The method we discuss in this paper combines several techniques mentioned above -- it uses the heterogeneous island model. The model runs a number of different stochastic optimization methods that periodically exchange some of the solutions they found. Apart from the optimization methods, there is also a planner that adaptively replaces the under-performing methods by better-performing ones and thus changes the set of algorithms used in the model online. The described system can thus be also considered an online parallel portfolio selection algorithm. From the point of view of evolutionary algorithms, the system is a way how to hybridize the evolutionary algorithm with other optimization methods in a general and modular way.

In this paper, we compare a number of different planners that decide which methods run in the heterogeneous island model at any point in time. The main goal of the paper is to provide a parallel system based on the idea that combines a number of general stochastic optimization methods in order to create a more general method with a better performance. In an ideal case, the performance of the combined method is better than the performance of each of the constituting methods separately, but even in case the performance is the same, the algorithm can be useful, if the same method can be run for multiple different types of problems. In such a case, it removes the need to select the correct algorithm for the problem at hand. The proposed method is evaluated on a wide range of problems including combinatorial optimization, continuous optimization and the hyper-parameter tuning of a machine-learning method. 

Some of the results presented in this paper have already be presented in a short paper~\cite{gecco2018}, namely the results of the P-QI and P-BM planners on the TSP and bin-packing problems. Here, we add eight new different planners, provide more detailed results of the baseline methods, and also present the results on seven new test instances from different fields -- vertex cover, continuous optimization, and hyper-parameter tuning in machine learning.

\section{Heterogeneous Island Model}

\begin{algorithm}[t]
\caption{General overview}
\label{alg:general_planner}
\begin{algorithmic}[1]
\State Initialize the methods uniformly on the islands
\State $t \gets 0$
\While{termination condition not met}
  \State $I \gets $ obtain information about running methods
  \If {there is a method $M$ that has not run}
    \State $k \gets $ the least useful method 
    \State $s \gets $ $M$
  \Else
    \State $k \gets $ select a method to kill using a planner-spec. rule
    \State $s \gets $ select a method to start using a planner-spec. rule
  \EndIf
  \State Kill method $k$ and start method $s$
  \State $t \gets t+1$
  \State Sleep until next planning iteration
\EndWhile
\end{algorithmic}
\end{algorithm}

In this section, we describe the heterogeneous island model with the re-planning of the methods. In the model, we assume a parallel computational environment with multiple CPU cores. Each CPU core corresponds to (and executes) a single island. Each island then executes a single optimization method. The methods running on different islands can be different, but they need to share the encoding of the candidate solutions as they share their solution among themselves.

During the run of the optimization, the system logs and processes information regarding the performance of various methods, like the quality of the solutions the methods share, how often the method improves the overall best solution or the number of distinct solutions it provided to other methods. The whole system is controller by a planner, that uses data measured during the execution of the methods in order to find a set of methods that would perform the best for the optimization problem at hand and for the current phase of the optimization. 

An overview of the system from the point of view of the planner is given in Algorithm~\ref{alg:general_planner}. The planner first initializes the island with different methods. Unless otherwise specified bellow (only for the random planner), the methods are assigned to the island in a round-robin manner, i.e. each method is executed the same amount of times (if possible). In case the number of different types of methods is larger than the number of islands, those methods that were not executed are given a preference to be executed first in re-planning.

After the initialization, the planner runs in a loop. In each iteration of the planner (planning iteration), the planner first obtains the information about the performance from the system and the methods themselves (line 4). If there is a method that has not been executed, an instance of the least useful method (according to planner philosophy, i.e. selected in the same way as on line 9 bellow) is removed and replaced by an instance of this method (lines 5-7, 12). Otherwise, the method that shall be removed and the method that shall be started in the given planning iteration are selected in a planner-specific way (lines 8-12). At the end of the planner iteration, the planner first removes the instance of the method that was selected to be removed and starts a new instance of the method selected to start. The planner then sleeps and waits for the start of the next iteration.

In the implementation, both the length of each planning iteration and the frequency of communication are time-based. While this makes the results of the experiments dependent on the hardware and on the implementation of the method (compared to an implementation that would be based on e.g. the number of function evaluations), the various optimization methods can run completely independently of the planner and the other methods in a fully asynchronous way and do not need to wait for the other methods to finish their evaluations. It also simplifies the implementation.

We also do not consider the parameters of the optimization methods in any way. While different settings of e.g. and evolutionary algorithm could be considered a different optimization method, it may actually make sense to work with the parameters more explicitly. This is however left for a future work.

\begin{algorithm}[t]
\caption{General Optimization Method}
\label{alg:general_method}
\begin{algorithmic}[1]
\State Create the initial (set of) solutions
\State $t \gets 0$
\While{termination condition not met}
  \State Generate new solution(s) based on the current state
  \State Receive solutions from the other methods in the system
  \State Incorporate received solution into own set of solutions
  \State Share the best solution
  \State Update information for the planners
\EndWhile
\end{algorithmic}
\end{algorithm}

\subsection{Modification of Methods}

As we already indicated, the common optimization methods must be slightly modified before they can be used in the heterogeneous island model. First of all, the methods must be able to share their best individuals with other methods present in the system, and they must also be able to receive individuals created by other methods and incorporate them into their optimization loop.

We assume the optimization methods used in the heterogeneous island model contain a main loop (e.g. the generational loop in evolutionary algorithm). In such a case, modification of the method is simple (cf. Algorithm~\ref{alg:general_method}) -- at the end of the optimization loop (line 5) new individuals are received from the other methods in the system and they are incorporated into the optimization (line 6). Then, the best individuals are shared with the rest of the system (line 7).

For all the methods we tested, the addition of the new solution is made in the same way as if the solution was generated by the methods. For example, in hill-climbing the solution is accepted if it is better than the current best solution for the method. In such a case, it replaces the solution of this methods and the method continues from it. In evolutionary algorithms, the received solutions are added to the population and it is up to the selection, if they survive or not. 

The generating of the initial (set of) solutions (line 1) and of the new candidate solutions (line 4) is completely method-specific and can range from simple random sampling in random search to a complex combination of various genetic operators in evolutionary algorithms.

Apart from the communication, the methods also must be able to provide information to the planner (line 8). Specifically, each method counts how many times other methods provided it with a better solution (for the helper planner described bellow). Each individual in the system also carries the history of methods that modified it in any way, and this information must be kept updated.

While the modifications (and especially the book-keeping of the information for planners) seem tedious, they can be implemented in a general way and most of the code can be shared by most of the methods. 

\subsection{Planners}

For the experiments in this paper, we designed a number of planners, that follow (with the exception of the random planner) the general template given in Algorithm~\ref{alg:general_planner}. All the planners first initialize all the islands with the available methods uniformly, i.e. each method is executed the same number of times. In case there are more methods than islands, the methods that have not been executed have a higher priority to be executed during re-planning. The planners also observe the system and in case new islands are added during the run of the algorithm, they initialize them with a method.

From the observation of the system, that planner obtains the following features that can be used during re-planning:
\begin{enumerate}
  \item quantity of improvement -- the number of times the given method improved the quality of the best solution,
  \item average fitness -- the average fitness of solutions shared by the method with other methods,
  \item quantity of material -- the number of distinct individuals each method shared with the rest of the methods,
  \item quality of material -- the number of times the method created a solution that is among the best $N$ solutions for a pre-defined $N$,
  \item helper number -- the number of times the method improved the quality of the current best solution of another method, and
  \item best solution contribution -- the number of times the given method was used in the history of the best solution.
\end{enumerate}

Features 1-3 can be computed directly by observing the solutions shared by the methods in the system. The helper number is computed by each of the methods and the planner can request this information. In order to compute the best solution contribution feature each individual contains its history that lists all the methods that were used to create it. 

In the rest of this section, we describe the details of the planners we propose in this paper. 

\begin{description}[labelindent=\parindent,leftmargin=0cm]
\item[Random (P-R)] planner is the simplest planner and also serves as a baseline. It initializes the methods randomly and in each planning iteration, it randomly eliminates one method and replaces it with a new randomly chosen method.
\item[Random with Guaranteed Chance (P-RG)] planner is a mo\-re sophisticated random planner -- it ensures each method is executed at least once and at the same time, it ensures that there are always at least $M_{min}$ different types of methods. In case the number of different running methods drops under the threshold, a random method is killed and replaced by the method that have not run recently. If there are methods that have not yet run, the planner chooses the method that has the least quantity of improvement and replaces it with a random method that has not run. Otherwise, the planner acts as the random planner. 
\item[Method Description (P-MD)] is based on the idea that in the initial phases of the optimization, exploration is more important than exploitation. Therefore, it divides the optimization methods into two sets -- exploitation and exploration ones. During the initialization, only exploration methods are used and during the computation, these are gradually replaced by the exploitation methods. Let $e$ be the number of exploration methods in the system, and $a$ be the number of all methods in the system. Let $t$ be the number of the current iteration and $T$ be the maximum number of iterations. If $ 1 - \frac{t}{T} < \frac{e}{a}$ then the exploration method that achieved the least quantity of improvement is killed and replaced by the exploitation method that achieved the best average of fitness so far. In the last iterations of the planner only exploration methods are used. Newly added method is protected and cannot be killed for the first $N_{protect}$ iterations. The division of the methods into the two groups is done manually.
\item[Best Helper (P-BH)] planner is a planner based on the number of times a method helped another method to improve its solution since the last iteration of the planner. A method $A$ helped another method $B$ if the solution received by $B$ from $A$ is better than the currently best solution of method $B$. In each planning iteration, the planner removes the method that helps the least and replaces is by the method, that helps the most. This planner does nothing in few first $N_{init}$ iterations, in order to wait for the performance of methods to stabilize.
\item[Best Average Fitness (P-AF)] planner uses the history of distributed computing and removes the method that has sent individuals with the worst average fitness during the last planing iteration, while the method with the best average fitness value of outgoing individuals is duplicated. Newly added methods are protected and ensured to run for at least $N_{protect}$ iterations. 
\item[Quantity of Improvement (P-QI)] planner uses the number of times each method produces a solution that is better than the current best one. During the initialization, methods are spread uniformly, and in each planning iteration, the method that produced the least number of improvements since the last re-planning is replaced by the method that produced the most improvements since the last re-planning. New methods are also protected for $N_{protect}$ planning iterations.
\item[Quantity Of Material (P-QM)] planner uses the information on the amount of distinct individuals sent by each method. During re-planning, the planner replaces the method that sent the smallest number of different individuals, by the method that created the largest number of different individuals. This planner also protects the newly started method for $N_{protect}$ iterations.
\item[Best Material (P-BM)] planner uses the information on the number of times the method provided a solution that is among the top $N$ solutions overall. In each planning iteration, the planner removes the method with the least number of top solutions since the last re-planning and replaces it with the method with the most top solutions.
\item[Best Solution Contribution (P-BC)] planner determines the importance of computational methods using the history of the operations applied to create the best individual. From this history, the planner uses the information about the methods that helped to create the best solution. In each re-planning iteration, the method that is most common in the history of the best individual replaces the method that contributed the least. The planner does nothing in the first $N_{init}$ iterations.
\item[Lazy Quantity Of Improvement (P-LQI)]
planner is similar to the quantity of improvement planner, but removes the methods only in case they are not beneficial, i.e. it replaces a method only in case it did not make any improvement to the best solution during the last $N_{patience}$ planning iterations. Such method is replaced with the one which achieved the greatest number of improvements during the last planner iteration. Newly added methods are protected for $N_{protect}$ iterations.
\end{description}

\section{Experiment settings}

In order to test the proposed portfolio selection algorithm, we ran a set of experiments where we compare the different planners to homogeneous island model running each of the optimization methods and to heterogeneous island model running all the methods, each in two instances, without re-planning (we also performed the same experiments with the single methods, which are equivalent to the homogeneous case without communication, but the results are generally worse than those with communication, therefore we do not present them here). The experiments are performed on five types of problems -- traveling salesman problem, bin packing problem, continuous optimization problem, vertex cover problem, and tuning of machine-learning hyper-parameters. These problems use a wide range of encodings and different genetic operators.

In all of the experiments, the parameters of the planners are set the same and they are given in Table \ref{tab:parameters}. We used these values of the parameters based on some preliminary experiments. Based on the settings, we can see that each run of the optimization takes approximately 50 minutes on 16 CPUs. The optimization methods share the individuals every 5 seconds. The island model uses a fully connected topology, i.e. each method can directly communicate with any other method. The runs are repeated nine times for each combination of a problem and a planner.

\begin{table}[t]
  \begin{center}
  \caption{The values of parameters shared by all the experiments (some of them are used by some planners only)}
    \label{tab:parameters}
    \begin{tabular}{l|l}
      \hline
      Parameter & Value \\
      \hline
      Number of iterations   & 50 \\
      Iteration length       & 60 s \\
      Number of runs         & 9 \\
      Number of islands      & 16 \\
      $N_{init}$             & 5 \\
      $N_{protect}$          & 3 \\
      $N_{patience}$         & 3 \\
      $M_{min}$              & 3 \\      
      \hline
    \end{tabular}
  \end{center}
\end{table}

We have implemented seven different optimization methods -- random search (RS), tabu search (TS), hill climbing (HC), simulated annealing (SA), evolutionary algorithm (EA), differential evolution (DE), and a brute force (BF) algorithm that performs a systematic search. All these share the following basic operators:
\begin{itemize}
  \item generate random solution -- used by random search and in initialization of EA and DE, 
  \item generate next solution -- used by the brute force method to generate the next solution systematically, 
  \item unary ``mutation'' operator -- used as a mutation in the EA and also by the hill climbing, simulated annealing and tabu search to generate the next solution, 
  \item binary ``crossover'' operator -- used by the EA as a crossover, and 
  \item ternary operator -- used in the DE by the differential mutation.
\end{itemize} 
There is no natural implementation of the last ternary operator in some cases, but we created an implementation nonetheless for the sake of completeness. We still call the resulting method ``differential evolution'', although the only feature common to this method and the differential evolution is the use of the ternary operator.
 
Some parameters of the optimization methods are also set the same for all the experiments, these are given in Table~\ref{tab:parametersOfMethods}. The other parameters depend on the particular optimization problem and are discussed in the rest of this section.

\begin{table}[t]
\caption{Parameters of the optimization methods}
\label{tab:parametersOfMethods}
\centering
	    \begin{tabular}{l|l|l}
	      \hline
	      Algorithm & Parameter & Value \\
	      \hline
	      Hill Climbing          & \# of neighbors & 10 \\
	      \hline
	      Random Search          &  $\emptyset$ \\
	      \hline
	      Evolution              & population size & 10 \\
	                             & mutation rate & 0.9 \\
	                             & crossover rate & 0.1 \\
	                             & selection & bin. tourn. \\
	      \hline
	      Brute Force            &  & \\
	      \hline
	      Tabu Search            & tabuModelSize & 50 \\
	      \hline
	      Simulated Annealing    & temperature&10,000 \\
	                             & coolingRate&0.002 \\
	      \hline
	      Differential Evolution & popSize & 50 \\
	                             & F & 1 \\  
	      \hline
	    \end{tabular}
\end{table}

\subsection{Traveling Salesman Problem}

The goal in the traveling salesman problem (TSP) is to find the shortest Hamiltonian cycle in a complete graph. For the experiments, we used two instances of the problem from the VLSI dataset\footnote{http://www.math.uwaterloo.ca/tsp/vlsi/}. One of them contains 1,083 cities (denoted as TSP1083 in the rest of this paper) and the other contains 662 cities (TSP662).

The TSP is a typical example of a problem, where the solutions are encoded as permutations. The permutation gives the order, in which the vertices of the graph should be visited.

In this case, the hill climbing, tabu search and simulated annealing use the 2-opt operator~\cite{Lin:1965} to generate new solutions. The random search method generates new permutations randomly and the brute force algorithm generates the next permutation in each step. The evolutionary algorithm uses the 2-opt operator as a mutation and it additionally uses a single-point crossover~\cite{Eiben:2015:IEC:2810085} -- a random crossover point is selected, the initial parts of the individuals are swapped and the rest of the individuals is filled by the rest of the numbers from the other parent in the order in which they appear in that parent. 

We also wanted to use the differential evolution and therefore we created a special ternary operator: first, the two permutations are subtracted and their difference is added to a third permutation. Then, for each value in the vector, we remember the index of the vector component on which it is located and the pairs $(value, index)$ are sorted by values. The sequence of indices then forms a new permutation. 

The optimization objective (and the values shown in the the results) is the length of the Hamiltonian cycle and should be minimized.

\subsection{Bin-packing Problem}

In the bin-packing problem (BPP), objects of various volumes are packed into bins of limited volume. The goal is to minimize the number of bins used. The solution of such problems is also commonly represented by a permutation that is decoded using the First Fit algorithm~\cite{princeton1971performance} in order to compute the number of objectives. For the experiments, we generated a random instance of the BPP with 1,000 objects with volumes between 0 and 1 and bins with the volume of 1. 

The random search, brute force and differential evolution methods generate the individuals in the same way as in the TSP problem. The hill climbing, tabu search, and simulated annealing use a displacement operator that moves a randomly selected number of consecutive values to the end of the permutation, the number of moved values is determined adaptively and is always less than 0.5 percent of all the values in the solution.

The evolutionary algorithm uses the order crossover~\cite{Eiben:2015:IEC:2810085} and a shift mutation that moves a random object to the end of the permutation.

The optimization objective is the number of bins and should be minimized.

\subsection{Continuous Optimization}

The goal of continuous optimization (CO) is to find the minimum of a function $f: \mathbb{R}^n \to \mathbb{R}$ in a multi-dimensional interval $[l_1, u_1] \times \dots \times [l_n, u_n]$. The solution is thus encoded as a vector of $n$ numbers from this interval. For the experiments in this paper we selected four functions from the BBOB benchmark\footnote{http://coco.gforge.inria.fr} -- the Büche-Rastrigin function (COf04), the Rosenbrock function (COf08), the Different Powers function (COf14) and the Schaffers function (COf17), all of them in 10-dimensional space. These specific functions were selected in order to evaluate the performance of the algorithm under different conditions regarding the multi-modality and separability of the functions. 

The tabu search, hill climbing, and simulated annealing methods generate the new individuals by adding a random number between -0.0025 and 0.0025 to each coordinate in the vector. The evolutionary algorithm uses the same operator as a mutation and additionally uses the weighted average of the coordinates of the vectors with weights randomly generated between 0 and 1. The random search generates random vectors from the whole multi-dimensional interval. The differential evolution uses the common implementation with parameters given in Table~\ref{tab:parametersOfMethods}. Finally, the brute force algorithm performs a grid search in the whole multi-dimensional interval -- it adds 0.005 to one of the coordinates in every step.

The optimization objective is directly the value of the function and should be minimized.

\subsection{Vertex Cover}

In the vertex cover (VC) problem, a graph is given and the goal is to find the smallest set of vertices of the graph, such that all the edges are covered. The solution is thus represented by a subset of vertices forming a valid coverage. The objective is to minimize the number of vertices in the cover. In this case, we used two benchmarks from the BHOSLIB\footnote{http://www.nlsde.buaa.edu.cn/\~kexu/benchmarks/graph-benchmarks.htm}, one with 1534 vertices (VCfrb59265) and the other with 4000 vertices (VCfrb10040).

The coverage is generally generated by starting from a smaller set of vertices and adding vertices until a coverage is created. In case of random search, the start is random and the vertices are also added randomly. In the hill climbing, tabu search and simulated annealing methods, five randomly selected vertices are removed and the cover is completed by adding (some of) their neighbors. The brute force algorithm generates all the subsets of the vertices and for subset that do not cover the whole graph, the cover is filled by adding random vertices. The evolutionary algorithm uses a mutation that removes three random vertices and finishes the cover by adding (some of) their neighbors. Finally, the differential evolution computes an intersection of two individuals and adds vertices from the third one until a valid cover is formed.

The optimization objective is the number of vertices in the cover and should be minimized.

\subsection{Hyper-parameter Tuning in ML}

The final type of problem we investigate in this paper is the tuning of the hyper-parameters of the machine learning methods (ML), i.e. we are searching for a set of hyper-parameters, such that the error rate of the model is minimized. To this end, we used the random forest model as implemented in the Weka framework~\cite{Hall:2009:WDM:1656274.1656278} and optimize its parameters for the wilt dataset from the UCI machine learning repository~\cite{Lichman:2013}. The goal of this dataset is a classification into two classes based on five attributes. The particular implementation of the random forest has the following parameters and we search their optimal values in the given range:  $P \in [20, 100]$, $K \in [1, 6]$, $V \in [0.0001, 0.5]$,  $U \in \{0, 1\}$, $B \in \{0, 1\}$, $depth \in [1, 20]$,  $I \in [20, 30]$, and  $batchsize \in [80, 120]$.

The tabu search, simulated annealing, and hill climbing use an operator that changes a value by adding $\pm 1$ in case the parameter is an integer one and a random number less than 0.005 if the parameter is a real number. The random search generates a random set of the parameters, while the brute force method performs a grid search of the parameters. Evolution uses the operator used in simulated annealing as a mutation and additionally uses a one-point crossover. As all the arguments are numerical, the differential evolution works in the classical sense, the values just need to be rounded for the integer parameters after the operator is applied.

The optimization objective is the error rate of the model and should be minimized.


\section{Results}

\begin{table*}[htp]
\caption{Results of the planners. The numbers represent the average value of the objective over nine independent runs. The numbers in subscript and superscript show the minimum and maximum value of the objective achieved in the experiments.}
\label{tab:results}
\centering
\def\arraystretch{1.25}
\begin{tabular}{ l|rrrrr }
\hline
Optimization Method & TSP1083 & TSP662 & BPP1000 & VCfrb59265 & VCfrb10040 \\
\hline
Brute force                & $104555_{103621}^{106075}$      & $50645_{49461}^{51293}$      & $638.70_{629}^{642}$     & $82.11_{77}^{87}$     & $142.11_{136}^{144}$ \\
Differential evolution     & $81407_{19903}^{100318}$        & $14573_{11343}^{26679}$      & $627.55_{626}^{629}$     & $43.11_{42}^{46}$     & $96.33_{88}^{100}$   \\
Evolution                  & $22271_{22043}^{22602}$         & $6686_{6504}^{6838}$         & $521.66_{520}^{524}$     & $44.11_{41}^{47}$     & $82.55_{78}^{89}$    \\
Hill Climbing              & $5154_{5084}^{5221}$            & $3044_{2968}^{3088}$         & $550.88_{545}^{555}$     & $35.00_{33}^{36}$     & $68.44_{65}^{72}$    \\
Random Search              & $100494_{100165}^{100845}$      & $48661_{48163}^{48968}$      & $628.77_{627}^{631}$     & $75.77_{74}^{78}$     & $141.44_{137}^{146}$ \\
Simulated annealing        & $14837_{14434}^{15280}$         & $7610_{7363}^{7972}$         & $564.88_{561}^{567}$     & $40.44_{38}^{45}$     & $80.44_{75}^{88}$    \\
Tabu search                & $5177_{5114}^{5239}$            & $2997_{2946}^{3073}$         & $549.88_{546}^{553}$     & $34.88_{34}^{36}$     & $64.55_{63}^{66}$    \\
\hline
Hetero - Static        	   & $5253_{5105}^{5313}$            & $3055_{2994}^{3102}$         & $529.22_{527}^{535}$     & $35.88_{34}^{38}$     & $70.11_{68}^{72}$ \\
\hline
Hetero - P-R               & $5378_{5281}^{5470}$            & $3051_{2945}^{3112}$         & $529.44_{527}^{536}$     & $36.22_{35}^{37}$     & $70.22_{66}^{78}$ \\
Hetero - P-RC      		   & $5386_{5228}^{5591}$            & $3048_{2981}^{3143}$         & $528.33_{526}^{533}$     & $36.22_{35}^{37}$     & $69.22_{68}^{73}$ \\
Hetero - P-MD     		   & $20494_{19769}^{21590}$         & $6521_{5764}^{6741}$         & $520.33_{518}^{523}$     & $36.11_{34}^{37}$     & $70.22_{68}^{74}$ \\
Hetero - P-LQI       	   & $5084_{5027}^{5189}$            & $3011_{2942}^{3119}$         & $523.66_{521}^{527}$     & $36.00_{34}^{37}$     & $69.66_{67}^{74}$ \\
Hetero - P-AF   		   & $5058_{5002}^{5126}$            & $3033_{2985}^{3089}$         & $528.30_{525}^{537}$     & $35.77_{34}^{38}$     & $68.88_{66}^{72}$ \\
Hetero - P-BH          	   & $5155_{5101}^{5284}$            & $3050_{2984}^{3137}$         & $525.44_{523}^{528}$     & $36.11_{35}^{38}$     & $69.77_{65}^{76}$ \\
Hetero - P-BM       	   & $4985_{4912}^{5035}$            & $3014_{2947}^{3062}$         & $523.22_{520}^{527}$     & $35.66_{33}^{38}$     & $70.00_{68}^{72}$ \\
Hetero - P-QI    		   & $5022_{4951}^{5098}$            & $3005_{2960}^{3080}$         & $525.22_{522}^{532}$     & $36.22_{35}^{37}$     & $69.11_{65}^{72}$ \\
Hetero - P-QM    		   & $6897_{6127}^{7684}$            & $3326_{3215}^{3441}$         & $526.66_{523}^{532}$     & $36.22_{34}^{38}$     & $71.11_{69}^{73}$ \\
Hetero - P-BC              & $5168_{5075}^{5271}$            & $3039_{2962}^{3110}$         & $525.66_{522}^{532}$     & $36.22_{35}^{38}$     & $68.44_{65}^{73}$ \\
\hline
\hline
Optimization method & COf04 & COf08 & COf14 & COf17 & MLWilt \\
\hline
Brute force                & $366.8_{164.0}^{946.2}$      & $35410_{7263}^{72908}$        & $17.79_{8.953}^{32.75}$   & $8.3948_{4.9845}^{10.792}$   & $0.0540_{0.0539}^{0.0539}$ \\
Differential evolution     & $5.104_{0.742}^{9.297}$      & $48.124_{5.5458}^{104.33}$    & $0.106_{0.036}^{0.182}$   & $1.3245_{0.7507}^{1.7501}$   & $0.0128_{0.0124}^{0.0132}$ \\
Evolution                  & $16.25_{10.94}^{24.87}$      & $0.0013_{0.0008}^{0.0015}$    & $0.000_{0.000}^{0.000}$   & $0.9105_{0.2749}^{1.9234}$   & $0.0128_{0.0124}^{0.0134}$ \\
Hill Climbing              & $187.4_{61.69}^{476.5}$      & $0.0011_{0.0007}^{0.0018}$    & $0.000_{0.000}^{0.000}$   & $5.4737_{4.9727}^{6.3656}$   & $0.0138_{0.0132}^{0.0143}$ \\
Random Search              & $91.065_{81.68}^{100.58}$    & $1782.9_{1416.9}^{2287.9}$    & $4.093_{3.376}^{5.145}$   & $3.7950_{3.1385}^{4.4831}$   & $0.0132_{0.0126}^{0.0134}$ \\
Simulated annealing        & $76.01_{19.91}^{147.3}$      & $5.3058_{0.1430}^{8.9798}$    & $0.000_{0.000}^{0.001}$   & $5.8375_{3.7791}^{8.7360}$   & $0.0135_{0.0126}^{0.0141}$ \\
Tabu search                & $208.5_{50.74}^{519.3}$      & $0.0012_{0.0009}^{0.0014}$    & $0.000_{0.000}^{0.000}$   & $5.3373_{4.4686}^{6.7744}$   & $0.0138_{0.0130}^{0.0143}$ \\
\hline
Hetero - Static            & $11.17_{5.970}^{24.87}$      & $0.0024_{0.0018}^{0.0029}$    & $0.000_{0.000}^{0.000}$   & $1.8958_{0.4755}^{2.8896}$   & $0.0128_{0.0124}^{0.0134}$ \\
\hline
Hetero - P-R               & $2.161_{0.000}^{3.980}$      & $0.0010_{0.0004}^{0.0020}$    & $0.000_{0.000}^{0.000}$   & $1.0786_{0.3701}^{1.6476}$   & $0.0128_{0.0124}^{0.0132}$ \\
Hetero - P-RG      	       & $2.211_{0.000}^{5.970}$      & $0.0013_{0.0002}^{0.0019}$    & $0.000_{0.000}^{0.000}$   & $1.0616_{0.2756}^{1.5171}$   & $0.0131_{0.0126}^{0.0134}$ \\
Hetero - P-MD       	   & $1.686_{0.000}^{13.93}$      & $0.0018_{0.0014}^{0.0022}$    & $0.000_{0.000}^{0.000}$   & $0.8883_{0.1674}^{1.5104}$   & $0.0128_{0.0124}^{0.0132}$ \\
Hetero - P-LQI       	   & $0.994_{0.000}^{2.985}$      & $0.0019_{0.0008}^{0.0026}$    & $0.000_{0.000}^{0.000}$   & $1.5419_{0.9240}^{2.7759}$   & $0.0130_{0.0124}^{0.0132}$ \\
Hetero - P-AF   	       & $1.106_{0.000}^{2.985}$      & $0.0026_{0.0018}^{0.0037}$    & $0.000_{0.000}^{0.000}$   & $1.2920_{0.4295}^{2.2167}$   & $0.0130_{0.0124}^{0.0132}$ \\
Hetero - P-BH          	   & $7.366_{0.000}^{23.58}$      & $0.0016_{0.0000}^{0.0032}$    & $0.000_{0.000}^{0.000}$   & $1.0468_{0.6861}^{1.3702}$   & $0.0128_{0.0124}^{0.0132}$ \\
Hetero - P-BM       	   & $0.884_{0.000}^{2.985}$      & $0.0026_{0.0018}^{0.0037}$    & $0.000_{0.000}^{0.000}$   & $0.9843_{0.4704}^{1.4008}$   & $0.0130_{0.0124}^{0.0132}$ \\
Hetero - P-QI    		   & $1.216_{0.000}^{2.985}$      & $0.0015_{0.0012}^{0.0019}$    & $0.000_{0.000}^{0.000}$   & $1.1269_{0.6168}^{1.8565}$   & $0.0129_{0.0124}^{0.0132}$ \\
Hetero - P-QM   		   & $19.08_{6.965}^{38.80}$      & $0.0022_{0.0006}^{0.0035}$    & $0.000_{0.000}^{0.000}$   & $1.7777_{0.9309}^{2.8130}$   & $0.0131_{0.0124}^{0.0134}$ \\
Hetero - P-BC              & $3.850_{0.000}^{14.92}$      & $0.0034_{0.0015}^{0.0054}$    & $0.000_{0.000}^{0.000}$   & $1.1631_{0.6877}^{1.5244}$   & $0.0129_{0.0124}^{0.0132}$ \\
\hline
\end{tabular}
\end{table*}

In order to run the experiments, all the above described methods were implemented in the JADE~\cite{bellifemine2007developing} multi-agent framework. It allows for simple implementation of distributed systems. Each of the islands corresponds to a single container running (apart from a few management agents) the optimization method, which is again implemented as another agent. The infrastructure provided by the JADE system makes the implementation of the communication and sharing in the system simple. 

The results of the experiments are provided in Table~\ref{tab:results}. It shows, for each of the problems and for each different setting of both homogeneous and heterogeneous islands the average objective value as well as the minimum and maximum objective over nine independent runs.

We can immediately see that, rather unsurprisingly, for the homogeneous models, different methods work better for different optimization problems. For the TSP and VC problems, the tabu search and hill climbing provide the best results, for the BPP, the best methods are tabu search and evolutionary algorithm. In continuous optimization, the differential evolution and evolutionary algorithm provide good results, and in some cases also hill climbing and tabu search work well. Finally, with the ML problem, the difference between the methods is rather small, still, evolution and differential evolution have the best average of the nine runs.

If we compare the static heterogeneous island where each method has two instances without re-planning, we can already see one of the advantages of the heterogeneous models -- the results are always close the results of the best homogeneous method, in some cases (COf04 and ML) the results are even slightly better. This means that even without re-planning it may make sense to run heterogeneous models instead of the homogeneous ones, in case multiple CPUs are available and parallel optimization is desired. The use of heterogeneous islands seems to remove the need to select the correct optimization algorithm and can be in fact considered an algorithm selection method.

The different planners significantly influence the results. In all cases the best planner (for given problem) is better than the static case without re-planning. In fact, the heterogeneous model with the P-BM (best genetic material) planner has at least the same performance as the static heterogeneous model in most cases (except COf08, ML). A similar observation holds for the P-QI (quantity of improvement) planner. 

The heterogeneous island model with re-planning also provides better results than the best homogeneous island model in six of the ten cases (one TSP benchmark, the bin-packing benchmark, and all four of the CO benchmarks). In the other cases, the homogeneous tabu search was better for the other TSP benchmark and for both the vertex cover benchmarks. The differential evolution and the evolutionary algorithms were better for the hyper-parameter tuning benchmark. However, in the latter case, the differences between the methods are negligible. In the cases where the homogeneous islands provided the best results, the difference between the homogeneous island and the best heterogeneous one were small. For example, in the TSP662 benchmark, the tabu search found on average solution of length 2997, while the P-QI found a solution of length 3005. The biggest difference between the homogeneous islands and the heterogeneous islands was observed on the VCfrb10040 benchmark, where the tabu search got a solution with average objective of 64.55 compared to the 68.44 for the best heterogeneous model.

\begin{table*}[t]
\caption{The number of times the given planner found a solution in the top quartile of the solutions found by all the methods.}
\label{tab:heteroQuartiles}
\centering
\begin{tabular}{ l|rr|r|rrrr|rr|r|r }
\hline
      & \multicolumn{2}{|c|}{TSP}    & BP    & \multicolumn{4}{|c|}{CO}         & \multicolumn{2}{|c|}{VCfrb}  & ML    & Sum \\
Planner                    & 1083 & 662                   & 1000   & f04   & f08       & f14     & f17     & 10040    & 59265             & Wilt  &     \\
\hline
 Static                     & 0     & 1                    & 0      & 0     & 0       & 1      & 2      & 0        & 1                 & 4     & 9   \\
 P-R                            & 0     & 2                    & 0      & 1     & 7       & 3      & 3      & 1        & 0                 & 1     & 18  \\
 P-RG               & 0     & 3                    & 0      & 3     & 4       & 2      & 3      & 0        & 0                 & 0     & 15  \\
 P-MD                & 0     & 0                    & 8      & 6     & 1       & 0      & 3      & 0        & 1                 & 3     & 22  \\
 P-LQI                & 4     & 5                    & 4      & 2     & 1       & 3      & 0      & 2        & 1                 & 1     & 23  \\
 P-AF            & 5     & 3                    & 0      & 4     & 0       & 3      & 1      & 3        & 2                 & 1     & 22  \\
 P-BH                   & 0     & 1                    & 0      & 1     & 3       & 4      & 4      & 2        & 0                 & 3     & 18  \\
 P-BM                & 9     & 4                    & 5      & 4     & 0       & 3      & 3      & 0        & 1                 & 1     & 30  \\
 P-QI             & 6     & 3                    & 2      & 2     & 4       & 5      & 3      & 3        & 0                 & 1     & 29  \\
 P-QM             & 0     & 0                    & 0      & 0     & 3       & 0      & 0      & 0        & 2                 & 1     & 6   \\
 P-BC                          & 0     & 2                    & 2      & 1     & 1       & 0      & 2      & 4        & 0                 & 2     & 14  \\
\hline
\end{tabular}
\end{table*}

In order to compare the various planners among themselves, we also computed how many times each planner found a results that is among the top results found overall. We define a results to be top if it is the top quartile of all the results found by all the methods. The results of this experiment are displayed in Table~\ref{tab:heteroQuartiles}. As we made nine independent runs, the best a planner can achieve is to have all the nine results among the top overall. This happened only once for the P-BM planner in the TSP1083 benchmark.
Overall, the P-BM planner and the P-QI planner provide the best results, with the former finding a top result in 30 cases and the latter providing such results in 29 cases. For comparison, the static heterogeneous islands found a top solution only in 9 runs. 

Interestingly, the random planners provide quite good results for the CO benchmarks, where in one case the P-R planner found solution in the top quartile in 7 of the 9 runs. On the other hand, the random methods failed in most of the other benchmarks. The P-MD planner based on the method description obtained good results in the BP benchmark and in the COf04 benchmark (with 8 and 6 top results respectively), but in the rest of the benchmark its performance was rather poor.  The P-BC planner is based on the history of the best individual and thus requires the most complex changes in the optimization methods. However, its performance is not very good (comparable to the random planners) and thus it does not seem to be worth the more complex implementation.

\section{Conclusion and Future Work}

We presented an algorithm for online portfolio selection. The presented model is a generalization of the homogeneous island model to cases, where each island runs a different optimization method. As such, it can be considered a way how to create hybrid optimization algorithms. Thanks to the modular implementation, methods can be hybridized easily and the only requirement is that they share the encoding of the solution.

We have shown that the heterogeneous island model provides better or comparable results compared to the best method executed in the homogeneous island model. It means that if multiple CPU cores should be used for optimization and the method needs to be selected, it is better to use the heterogeneous model that selects the method automatically than using the homogeneous model and selecting the best method in multiple runs. As such, the heterogeneous models can also be consider and algorithm selection method without any interaction with the user.

So far, we have experimented with only seven different optimization methods with fixed parameters. However, in principle, we could used much more methods or different sets of their parameters. While methods with different parameters could be considered a completely different method and used in the same way as described in this paper. It may make sense to define a similarity between the methods and take it into account during the planning. Such extensions are left for future work.

\section*{Acknowledgment}

Martin Pilát has been supported by Czech Science Foundation project number 17-17125Y. Štěpán Balcar has been supported by SVV project number SVV-2018-260451.

\bibliographystyle{unsrt}
\bibliography{literature}

\end{document}